\documentclass[a4paper, 10pt, conference]{ieeeconf}      
\usepackage{graphicx}
\usepackage{amsfonts}
\usepackage{csquotes}
\IEEEoverridecommandlockouts                              

\title{\LARGE \bf
Speed-up of Self-Organizing Networks for Routing Problems in a Polygonal Domain
}

\author{Miroslav Kulich, Roman Sushkov, Libor P\v{r}eu\v{c}il
\thanks{The authors are with Czech Institute of Informatics, Robotics and Cybernetics, Czech Technical University in Prague,%
      Zikova 1903/4, 166 36 Prague, Czech Republic%
   {\tt\small ~~e-mail: kulich@cvut.cz, sushkrom@fel.cvut.cz, preucil@cvut.cz}\newline%
\indent This work has been supported by the European Union's Horizon 2020 research and innovation programme under grant agreement No 688117 and the Technology Agency of the Czech Republic under the project no.~TE01020197 \enquote{Centre for Applied Cybernetics}.
}
}

\begin{document}

\maketitle
\thispagestyle{empty}
\pagestyle{empty}

\begin{abstract}
Routing problems are  optimization problems that consider a set of goals in a graph to be visited by a vehicle (or a fleet of them) in an optimal way, while numerous constraints have to be satisfied. 
We present a solution based on multidimensional scaling which significantly reduces computational time of a self-organizing neural network solving a typical routing problem -- the Travelling Salesman Problem (TSP) in a polygonal domain, i.e. in a space where obstacles are represented by polygons. 
The preliminary results show feasibility of the proposed approach and although the results are presented only for TSP, the method is general so it can be used also for other variants of routing problems.   

\end{abstract}

\section{INTRODUCTION}

Mobile robots, as their autonomy increases, can be employed in more and more complex scenarios of e.g. inspection, reconnaissance or surveillance. 
A mission goal in these scenarios is not specified in the form 'go from A to B'. 
Instead, the task incorporates specification of particular sub-goals to be reached by a robot successively as well as determination of the order in which to reach the sub-goals.
Finding an optimal solution leads frequently to some kind of combinatorial optimization problems like the TSP, the Vehicle Routing Problem or their variants called routing problems, which are NP-hard in general.

Although powerful heuristics for TSP providing near optimal solutions in a reasonable time exist, they have several drawbacks.
These heuristics are not general so they cannot be used for other variants of problems, for multi-robot cases, and incorporation of additional constrains (e.g. assuming heterogeneous robots) is not straightforward or even possible.
Therefore, other approaches like genetic algorithms, ant colony optimization, integer linear programming, self-organizing maps and others have an important role in situations where heuristics can not be applied. 

A self-organizing map (SOM) as a popular approach to routing problems~\cite{faigl2011application,Faigl2014,Cochrane}
is a competitive learning network trained using iterative unsupervised learning. 
Assuming the TSP as the problem of finding the shortest closed path between a defined set of goals in 2D, a standard variant of SOM is expressed as a ring of neurons where two neighboring neurons are connected and neurons are represented by their coordinates in a plane. 
The learning procedure takes one input goal in each iteration and finds the nearest neuron to it. 
Positions of all neurons are then adapted based on their ring distance to the winning neuron (i.e. the number of neurons between the neuron and the winning neuron), iteration number, and possibly other criteria.
The process finishes when some stopping condition (e.g., a number of iterations) is satisfied.     

This approach works fast in an empty space, where goal-neuron distance needed to determine the nearest neuron is determined as the Euclidean distance. 
The situation is more complicated in the real world when obstacles are present, as a shortest collision-free path between a goal and a neuron has to be computed, which is a computationally intensive task.

We presented several approaches to fast determination of a neuron-goal path in our previous work considering a polygonal representation of obstacles which are based on shortest path maps~\cite{kulich05ssrr} or convex polygon partitioning~\cite{faigl2011application}. 
The idea presented in this paper is different: a polygonal domain representing the working environment with obstacles is transformed into a higher-dimensional space without obstacles so that distances in this space are (ideally) the same as those in the input space.

\begin{table*}[t]
\centering
\begin{tabular}{ r r r | r r r r r | r r r}
\hline
Problem & $n$ & $L_{opt}$ & \multicolumn{5}{c}{Transformation based}  & \multicolumn{3}{|c}{Shortest Paths (va-10)} \\
 & & & PDM & PDB & $T$(s) & $T_{MDS}$(s)& $T_{SOM}$(s)  & PDM & PDB & $T$(s) \\
\hline
$jari^*$ & 6 & 13.6 & 4.41 & 1.47 & 0.01 & 0.0 & 0.0 & 0.21 & 0.0 & 0.08\\
$complex2$ & 8 & 58.5 & 0.0 & 0.0 & 0.03 & 0.02 & 0.0 & 0.29 & 0.0 & 0.1\\
$m1^*$ & 13 & 17.1 & 0.0 & 0.0 & 0.01 & 0.0 & 0.0 & 0.14 & 0.0 & 0.16\\
$m2$ & 14 & 19.4 & 9.79 & 9.28 & 0.03 & 0.02 & 0.01 & 12.51 & 8.95 & 0.17\\
$map^*$ & 17 & 26.5 & 5.66 & 5.28 & 0.02 & 0.0 & 0.01 & 3.85 & 0.0 & 0.25\\
$potholes$ & 17 & 88.5 & 1.58 & 0.0 & 0.37 & 0.28 & 0.01 & 1.48 & 0.0 & 0.33\\
$a^*$ & 22 & 52.7 & 0.19 & 0.0 & 0.03 & 0.0 & 0.01 & 0.28 & 0.0 & 0.4\\
$rooms$ & 22 & 165.9 & 6.21 & 4.04 & 0.09 & 0.06 & 0.02 & 1.01 & 0.17 & 0.44\\
$dense_4^*$ & 53 & 179.1 & 14.35 & 7.98 & 0.18 & 0.03 & 0.07 & 14.14 & 6.86 & 2.71\\
$potholes_2$ & 68 & 154.5 & 5.05 & 3.24 & 0.56 & 0.28 & 0.15 & 5.06 & 3.09 & 3.54\\
$jh_2$ & 80 & 201.9 & 5.45 & 1.98 & 0.68 & 0.35 & 0.21 & 2.04 & 0.43 & 4.96\\
$pb_4$ & 104 & 654.6 & 3.67 & 1.56 & 0.44 & 0.04 & 0.35 & 1.1 & 0.03 & 6.88\\
$ta_2^*$ & 141 & 328.0 & 17.01 & 13.26 & 0.86 & 0.37 & 0.44 & 2.96 & 2.13 & 11.94\\
$h2_5^*$ & 168 & 943.0 & 12.26 & 9.05 & 2.71 & 1.04 & 0.61 & 1.7 & 1.16 & 67.68\\
$potholes_1$ & 282 & 277.3 & 8.26 & 6.02 & 3.44 & 0.26 & 2.44 & 6.54 & 3.75 & 64.6\\
$pb_{1.5}$ & 415 & 839.6 & 8.83 & 7.36 & 5.62 & 0.04 & 5.17 & 2.17 & 1.07 & 123.92\\
$h2_2^*$ & 568 & 1316.2 & 28.66 & 24.94 & 24.81 & 13.47 & 7.23 & 2.38 & 1.75 & 686.24\\
$ta_1^*$ & 574 & 541.1 & 18.85 & 14.64 & 15.42 & 6.98 & 7.33 & 5.46 & 4.77 & 225.98\\
\hline
\end{tabular}
\caption{Comparison of TBA and the approach in~\cite{faigl2011application}
The quality of solutions is evaluated as the percent deviation to the
optimum tour length of the mean solution value, $PDM=(L-L_{opt})/L_{opt}\cdot 100\%$, and as the percent deviation from the optimum of the
best solution value (PDB), where $L_opt$ is the length of the optimal solution found by the Concorde solver, see~\cite{faigl2011application}.
Dimension of the transformed space is $m=5$ or $m=10$ (those are marked by an asterisk).} 
\label{table}
\end{table*}

\section{Algorithm}
Assume a polygonal workspace ${\cal W} \subset {\mathbb{R}^2}$ and a set of goals $\cal G$.
The task is to find a shortest path visiting all goals from $\cal G$ while respecting obstacles of $\cal W$.
The proposed Transformation Based Algorithm (TBA) works in the following steps:
\begin{enumerate}
\item Compute a distance matrix for all vertices of obstacles in $\cal W$. To do this, a visibility graph of the vertices is constructed on which  Johnson's algorithm to find the shortest paths between all pairs of vertices is run. 
\item Use multidimensional scaling to transform the vertices into $\mathbb{S}^m$, a high-dimensional space without obstacles, where their mutual distances are preserved.
\item Compute constrained Delaunay triangulation (CDT) of the original polygonal vertices. 
\item Approximate coordinates of the goals. This is done in several steps:
\begin{enumerate}
\item Determine a triangle of CDT, the goal lies in. 
\item Find a linear combination of vertices of the triangle  which represents the goal in ${\cal W}$.
\item Apply the same linear combination to determine a position of the goal in $\mathbb{S}^m$. 
\end{enumerate}
\item Run SOM in $\mathbb{S}^m$ (neuron-goal distances are computed as Euclidean distances in  $\mathbb{S}^m$).
\item Use the solution (ordering of the goals) found $\mathbb{S}^m$ as the results of the problem in the original space.
\end{enumerate}

The key component of the algorithm is multidimensional scaling (MDS)~\cite{borg2005modern} -- a method that maps a set of samples in some space into some other space based on their similarity which is usually specified by a positive symmetric distance matrix.
MDS is traditionally used for visualization of multidimensional data in two or three-dimensional space (or, mapping multidimensional data into low-dimensional space). 
In the produced output, similar objects are mapped close to each other forming clusters.
Contrary to the traditional use, the objects (vertices of obstacles) are mapped into a high-dimensional space in our case as our goal is to preserve the distances between the objects while abolishing the obstacles and not to visualise their similarity. 

\section{Preliminary results}
To demonstrate feasibility of the proposed approach, experiments were performed with  Scaling by Maximizing a Convex Function~\cite{borg2005modern} employed for MDS and ORC-SOM~\cite{zhang2012overall} as a variant of SOM and results were compared with those in ~\cite{faigl2011application}, which is one of the best SOM-based approaches.   
The preliminary results presented in Table~\ref{table} show that TBA is by one or two orders faster, while quality of generated solutions is slightly worse. 
Note also that MDS-based transformation does not depend on goals as it is computed from vertices of obstacles only. 
This implies that it can be precomputed once a map is available and $T_{SOM}$ is thus relevant if several TSP problems are to be solved for a given map.

\begin{figure}
\centering
\includegraphics[height=0.13\textheight]{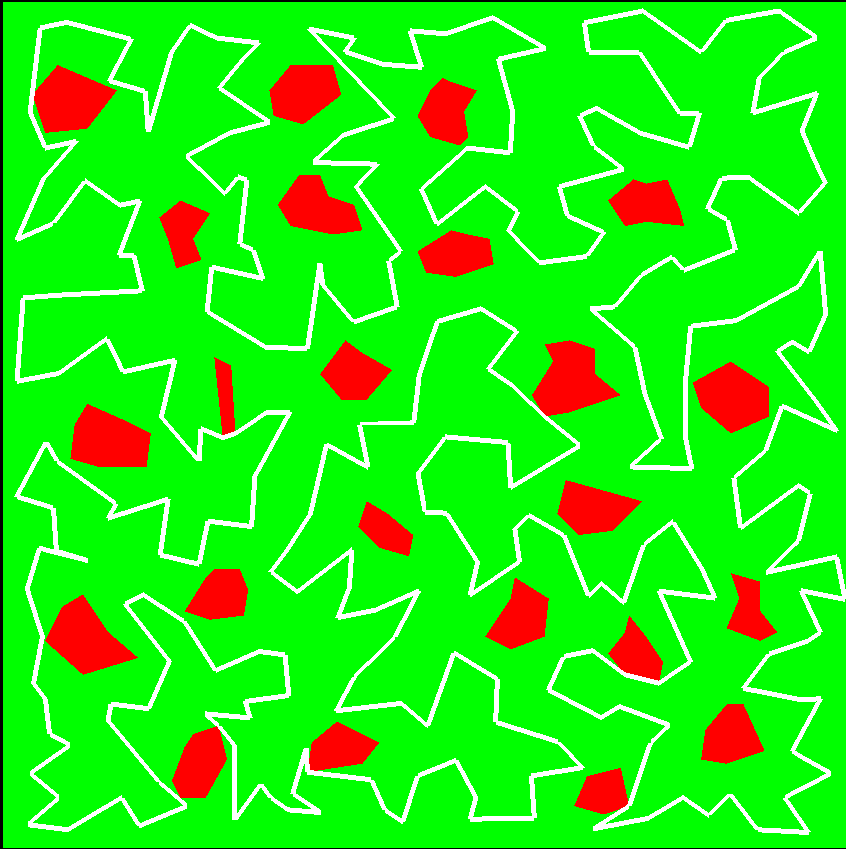}
\includegraphics[height=0.13\textheight]{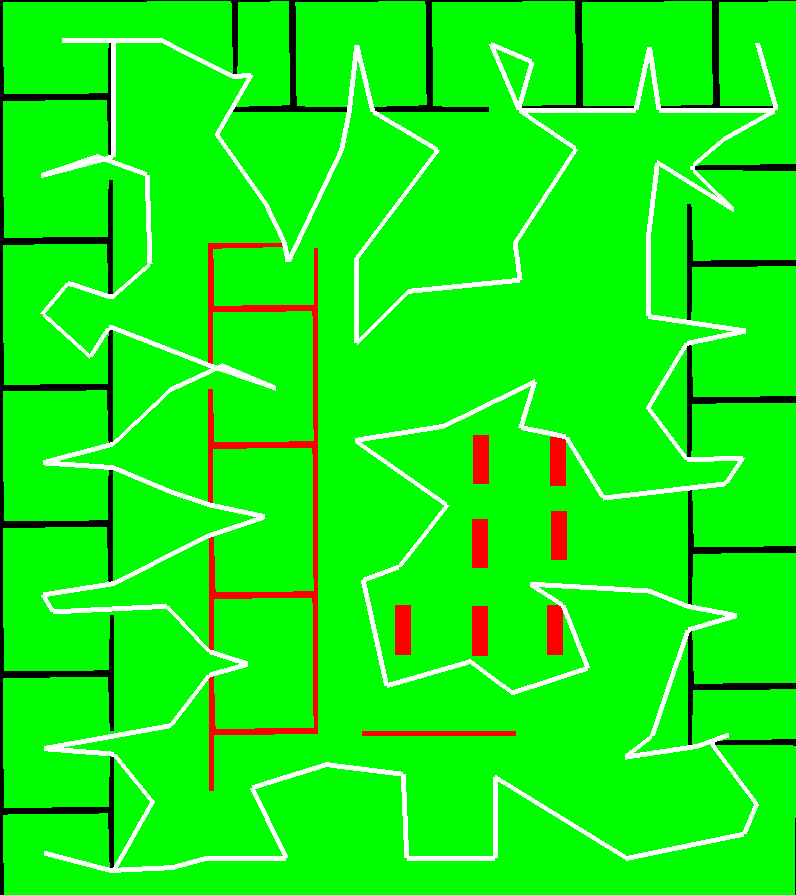}

\smallskip
\centering
\includegraphics[height=0.13\textheight]{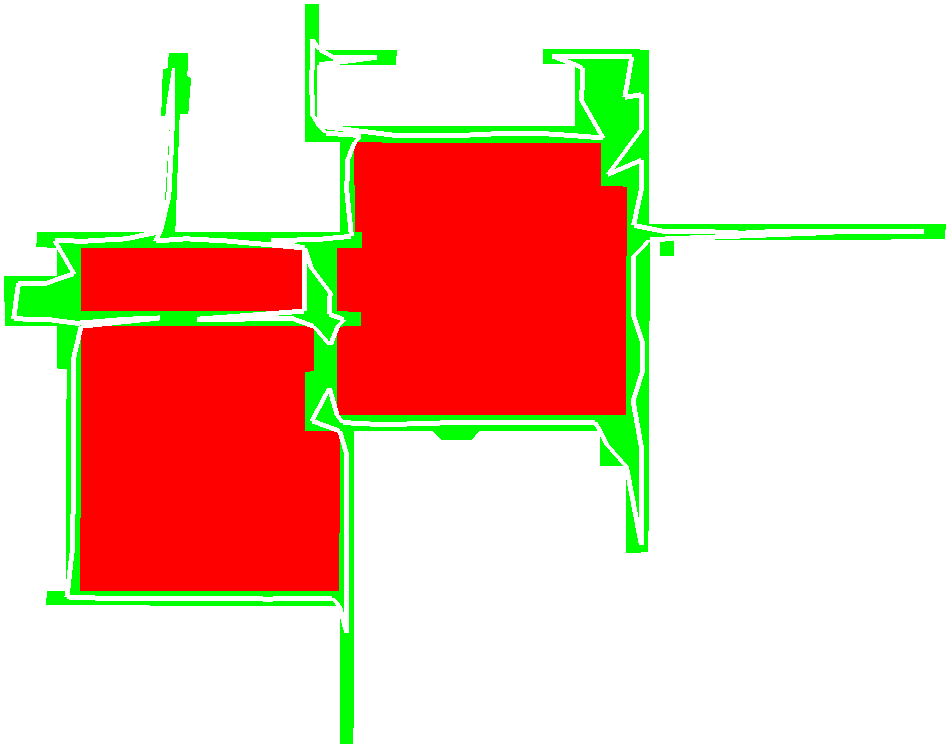}
\includegraphics[height=0.13\textheight]{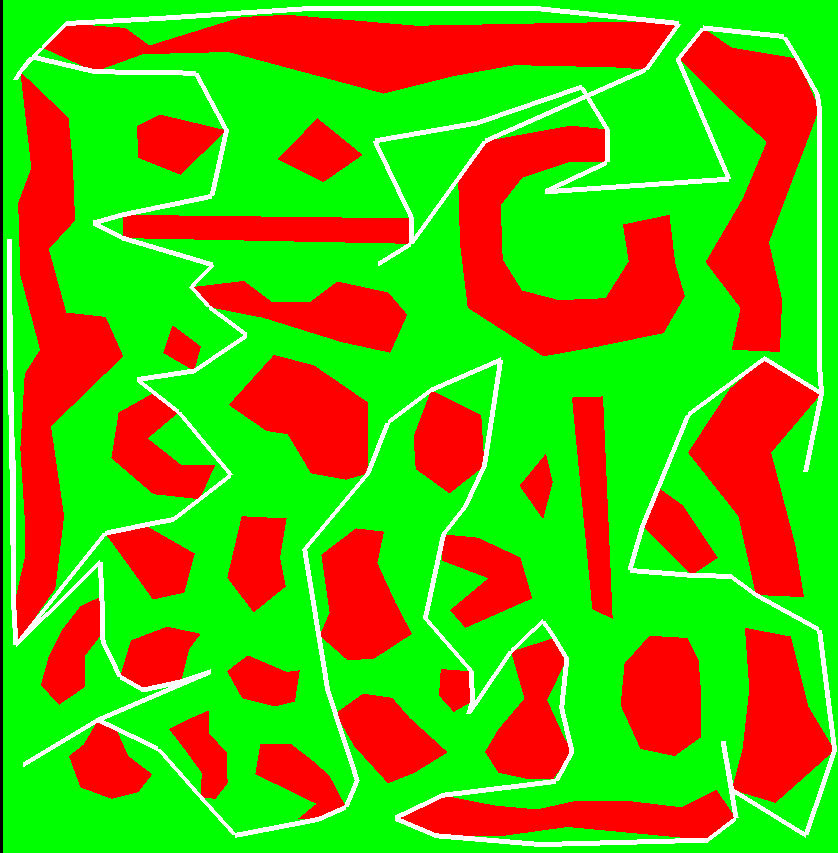}
\caption{Examples of solutions of some problems (goals are placed in all turning points of the polylines.}
\end{figure}

The current and future research is focused on study of properties of MDS and a polygonal domain in order to find better approximations of the polygonal domain and thus produce high-quality results.


\bibliographystyle{IEEEtran}
\bibliography{main}

\begin{thebibliography}{1}
\providecommand{\url}[1]{#1}
\csname url@rmstyle\endcsname
\providecommand{\newblock}{\relax}
\providecommand{\bibinfo}[2]{#2}
\providecommand\BIBentrySTDinterwordspacing{\spaceskip=0pt\relax}
\providecommand\BIBentryALTinterwordstretchfactor{4}
\providecommand\BIBentryALTinterwordspacing{\spaceskip=\fontdimen2\font plus
\BIBentryALTinterwordstretchfactor\fontdimen3\font minus
  \fontdimen4\font\relax}
\providecommand\BIBforeignlanguage[2]{{%
\expandafter\ifx\csname l@#1\endcsname\relax
\typeout{** WARNING: IEEEtran.bst: No hyphenation pattern has been}%
\typeout{** loaded for the language `#1'. Using the pattern for}%
\typeout{** the default language instead.}%
\else
\language=\csname l@#1\endcsname
\fi
#2}}

\bibitem{faigl2011application}
J.~Faigl, M.~Kulich, V.~Von\'{a}sek, and L.~P\v{r}eu\v{c}il, ``{An application
  of the self-organizing map in the non-Euclidean Traveling Salesman
  Problem},'' \emph{Neurocomputing}, vol.~74, no.~5, pp. 671--679, 2011.

\bibitem{Faigl2014}
J.~Faigl and G.~A. Hollinger, \emph{Self-Organizing Map for the
  Prize-Collecting Traveling Salesman Problem}.\hskip 1em plus 0.5em minus
  0.4em\relax Cham: Springer International Publishing, 2014, pp. 281--291.

\bibitem{Cochrane}
E.~Cochrane and J.~Beasley, ``The co-adaptive neural network approach to the
  euclidean travelling salesman problem,'' \emph{Neural Networks}, vol.~16,
  no.~10, pp. 1499 -- 1525, 2003.

\bibitem{kulich05ssrr}
M.~Kulich, J.~Faigl, and L.~P\v{r}eu\v{c}il, ``Cooperative planning for
  heterogeneous teams in rescue operations,'' in \emph{IEEE International
  Safety, Security and Rescue Rototics, Workshop, 2005.}, June 2005, pp.
  230--235.

\bibitem{borg2005modern}
I.~Borg and P.~J. Groenen, \emph{Modern multidimensional scaling: Theory and
  applications}.\hskip 1em plus 0.5em minus 0.4em\relax Springer Science \&
  Business Media, 2005.

\bibitem{zhang2012overall}
J.~Zhang, X.~Feng, B.~Zhou, and D.~Ren, ``An overall-regional competitive
  self-organizing map neural network for the euclidean traveling salesman
  problem,'' \emph{Neurocomputing}, vol.~89, pp. 1--11, 2012.

\end{thebibliography}

\end{document}